\documentclass[runningheads]{llncs}
\usepackage{graphicx}
\usepackage{amsmath,amssymb} 
\usepackage{color}
\usepackage[width=122mm,left=12mm,paperwidth=146mm,height=193mm,top=12mm,paperheight=217mm]{geometry}
\begin{document}
\pagestyle{headings}
\mainmatter

\title{A Large Contextual Dataset for Classification, Detection and Counting of Cars with Deep Learning.} 

\titlerunning{Classification, Detection and Counting of Cars with Deep Learning.}

\authorrunning{Mundhenk, Konjevod, Sakla and Boakye}

\author{T. Nathan Mundhenk, Goran Konjevod, Wesam A. Sakla, Kofi Boakye}


\institute{Computational Engineering Division,\\
        	Lawrence Livermore National Laboratory\\
        	\email{ \{mundhenk1,konjevod1,sakla1\}@llnl.gov, kaboakye@gmail.com}
}

\maketitle

\begin{abstract}
We have created a large diverse set of cars from overhead images\footnote{Data sets, annotations, networks and scripts are available from http://gdo-datasci.ucllnl.org/cowc/}, which are useful for training a deep learner to binary classify, detect and count them. The dataset and all related material will be made {\it publically available}. The set contains contextual matter to aid in identification of difficult targets. We demonstrate classification and detection on this dataset using a neural network we call {\it ResCeption}. This network combines residual learning with Inception-style layers and is used to {\it count cars in one look}. This is a new way to count objects rather than by localization or density estimation. It is fairly accurate, fast and easy to implement. Additionally, the counting method is not car or scene specific. It would be easy to train this method to count other kinds of objects and counting over new scenes requires no extra set up or assumptions about object locations. 
\end{abstract}

\section{Introduction}
Automated analytics involving detection, tracking and counting of automobiles from satellite or aerial platform are useful for both commercial and government purposes. For instance, \cite{OrbitalInsight16} have developed a product to count cars in parking lots for investment customers who wish to monitor the business volume of retailers. Governments can also use tracking and counting data to monitor volume and pattern of traffic as well as volume of parking. If satellite data is cheap and plentiful enough, then it can be more cost effective than embedding sensors in the road. 

A problem encountered when trying to create automated systems for these purposes is a {\it lack of large standardized public datasets}. For instance {\it OIRDS} \cite{OIRDS} has only 180 unique cars. A newer set {\it VEDAI} \cite{VEDAI} has 2950 cars. However, both of these datasets are limited by not only the number of unique objects, but they also tend to cover the same region or use the same sensors. For instance, all images in the VEDAI set come from the {\it AGRC} Utah image collection \cite{UtahData}. 

We have created a new large dataset of {\it Cars Overhead with Context} (COWC). Our set contains a large number of {\it unique cars} (32,716) from six different image sets each covering a different geographical location and produced by different imagers. The images cover regions from {\it Toronto Canada} \cite{TorontoData}, {\it Selwyn New Zealand} \cite{SelwynData}, {\it Potsdam} \cite{PotsdamData} and {\it Vaihingen Germany} \cite{VaihingenData}, {\it Columbus} \cite{ColumbusData} and {\it Utah} \cite{UtahData} {\it United States}.  The set is also designed to be difficult. It contains 58,247 usable negative targets. Many of these have been hand picked from items easy to mistake for cars. Examples of these are boats, trailers, bushes and A/C units. To compensate for the added difficulty, context is included around targets. Context can help tell us something may not be a car (is sitting in a pond?) or confirm it is a car (between other cars, on a road). In general, the idea is to allow a deep learner to determine the weight between context and appearance such that something that looks very much like a car is detected even if it's in an unusual place.

\section{Related Work}
We will focus on {\it three tasks} with our data set. The first task is a {\it two-class classifier}. To some extent, this is becoming trivial. For instance, \cite{Chen13} reports near 100\% classification on their set. This is part of the reason for trying to increase the difficulty of targets. Our contribution in this task is to demonstrate good classification on an intentionally difficult dataset. Also, we show that context does help with this task, but probably mostly on special difficult cases.  

A more difficult problem is {\it detection and localization}. A very large number of detectors start with a trained classifier and some method for testing spatial locations to determine if a target is present. Many approaches use less contemporary SVM and Boosting based methods, but apply contextual assistance such as road patch detection or motion to reduce false positives \cite{VEDAI,Moranduzzo14,Holt09,Kamenetsky15}. Some methods use a deep learning network with strided locations \cite{OrbitalInsight16,Chen13}  that generate a heat map. Our method for detection is similar to these, but we include context by expanding the region to be inspected in each stride. We also use a more recent neural network which can in theory handle said context better. 

By far our most interesting contribution that uses our new data set is vehicle counting. Most contemporary counting methods can be broadly categorized as a {\it density estimator} \cite{Zhang15,Arteta14,Lempitsky10} or, {\it detection instance counter} \cite{Moranduzzo14,Kamenetsky15,French15}. Density estimators try to create an estimation of the density of a countable object and then integrate over that density. They tend not to require many training samples, but are usually constrained to the same scene on which it was trained. Detection counters work in the more intuitive fashion of localizing each car uniquely and then counting the localizations. This can have the downside that the entire image needs to be inspected pixel by pixel to create the localizations. Also, occlusions and overlapping objects can create a special challenge since a detector may merge overlapping objects.  Another approach tries to count large crowds of people by taking a fusion over many kinds of feature counts using a Markov random field constraint \cite{Idrees13} and seems like a synthesis of the density and detection approaches.  However, it uses object-specific localizers such as a head detector so it is unclear how well it would generalize to other objects. 

Our method uses another approach. We teach a deep learning neural network to recognize the number of cars in an extended patch. It is trained only to count the number of objects {\it as a class} and is {\it not} given information about location or expected features. Then we count all the cars in a scene using a {\it very large stride} by counting them in groups at each stride location. This allows us to take one look at a large location and count by appearance. It has recently been demonstrated that {\it one-look} methods can excel at both speed and accuracy \cite{YOLO} for recognition and localization. The idea of using a one-look network counter to “learn to count” has recently been demonstrated on synthetic data patches \cite{Segui15} and by regression on subsampled crowd patches \cite{Wang15}. Here we utilize a more robust network, and demonstrate that a large strided scan can be used to {\it quickly count} a very large scene with reasonable accuracy. Additionally, we are not constrained by scene or location. Cars can be automatically counted {\it anywhere} in the world, even if they are not on roads or moving. 

\section{Data Set Details}
Overhead imagery from the six sources is standardized to 15 cm per pixel at ground level from their original resolutions. This makes cars range in size from 24 to 48 pixels. Two of the sets (Vaihingen, Columbus) are grayscale. The other four are in RGB color. Typically, we can determine the approximate scale at ground level from imagery in the field (given metadata from camera and GPS, IMU calibrated SFM \cite{Kelly09} or {\it a priori} known position for satellites). So we do not need to deal with scale invariance. However, we cannot assume as much in terms of quality, appearance or rotation. Many sets can still be in grayscale or have a variety of artifacts. Most of our data have some sort of {\it orthorectification artifacts} in places. These are common enough in most overhead data sets that they should be addressed here.

The image set is {\it annotated by single pixel points}. All cars in the annotated images have a dot placed on their center. Cars that have occlusions are included so long as the annotator is reasonably sure the item is a car. Large trucks are completely omitted since it can be unclear when something stops being a light vehicle and starts to become a truck. Vans and pickups are included as cars even if they are large. All boats, trailers and construction vehicles are always added as negatives. Each annotated image is methodically searched for any item that might at least slightly look like a car. These are then added. It is critical to try and include as many possible confounders as we can.  If we do not, a trained system will underperform when introduced to new data. 

Occasionally, some cars are highly ambiguous or may be distorted in the original image. Whether to include these in the patch sets depends on the task. For the {\it classification task}, if it was unclear if an item was or was not a car, it was left out. Distorted cars were included so long as the distortion was not too grave. In both cases, this is a judgment call. For the {\it counting task}, one is forced to deal with these items since they appear incidentally in many training patches. For that, a best guess is made. If a car was highly distorted, it was counted as a car so long as it appeared to be a car. 

\begin{figure}
\centering
\includegraphics[height=4.5cm]{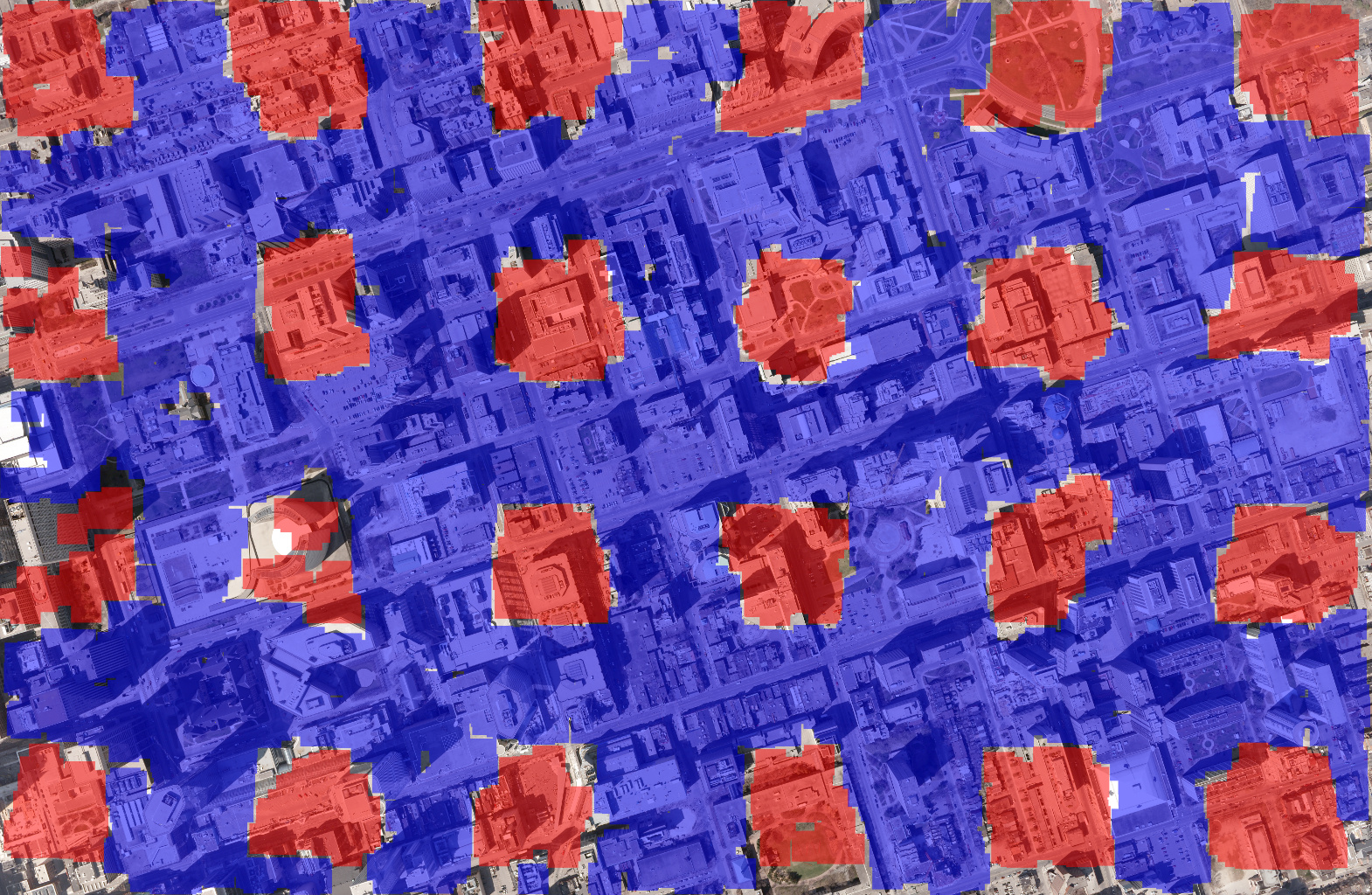}
\caption{The locations from which testing and training patches were extracted from an overhead image of Toronto. Blue areas are training patch areas while red areas are testing patch areas. }
\label{fig:patch_regions}
\end{figure}

To extract training and testing patches from large overhead images, they were subdivided into grids of size 1024x1024 (see Fig.~\ref{fig:patch_regions}). These grid regions were automatically assigned as training or testing regions. This keeps training and testing patches separate. The two types of patches cannot overlap. However, testing image patches may overlap other testing image patches and training patches overlap other training patches. In places like crowded parking lots, patches necessarily overlap. Every fourth grid location was used for testing. This creates an approximate ratio of more than three training patches for each testing patch. The patches are extracted in slightly different ways for different tasks. We do not include a validation set because we use a {\it held out set} of 2048x2048 scene images for final testing in the wild for each given task.  

\begin{figure}
\centering
\includegraphics[height=4.0 cm]{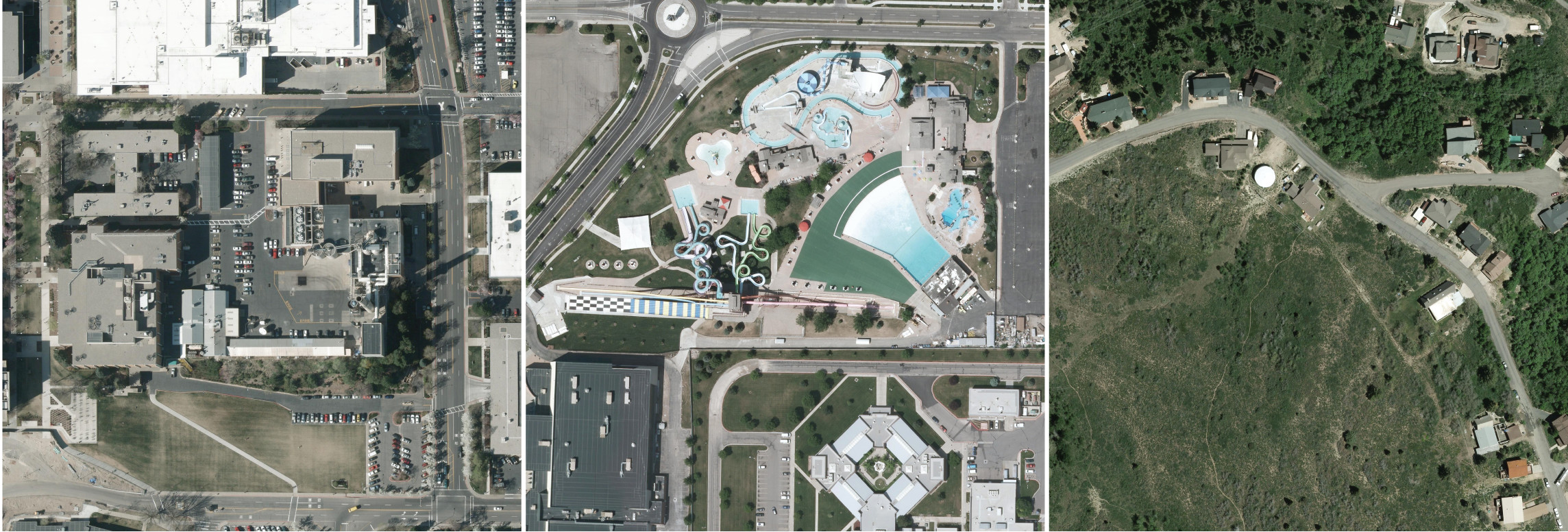}
\caption{Three examples of 2048x2048 held out scenes we used. These include a mixed commercial industrial area, a water park and a mountain forest area. }
\label{fig:scenes}
\end{figure}

Each held out scene is 2048x2048 (see Fig.~\ref{fig:scenes}). This is approximately 307x307 meters in size at ground level. Held out scenes are designed to be varied and {\it non-trivial}. For instance, one contains an auto reclamation junkyard filled with banged up cars. We have 10 labeled held out scene images, and 10 more where cars have been counted but not labeled. An additional 2048x2048 validation scene was used to adjust parameters before running on the held out scene data. The held out data is taken from the Utah set since there is an abundance of free data. They are also taken from locations far from where the patch data was taken. Essentially, all patch data was taken from {\it Salt Lake City} and all held out data was taken from {\it outside} of that metropolitan area. The held out data also contains mountainous wooded areas and a water park not found anywhere in the patch data. Other unique areas such as the aforementioned auto junkyard and a utility plant are included, but there are analogs to these in the patch data. 

\section{Classification and Detection}
We created a contextual set of patches for classification training. These are sized 256x256. We created rotational variants with 15 degree aligned offsets of each unique car and each unique negative. This yielded a set of 308,988 {\it training patches} and 79,447 {\it testing patches}. A patch was considered to contain a car if it appeared in a central 48x48 region (The largest expected car length). Any car outside this central region was considered context. So, negative patches frequently had cars in them, so long as the car did not fall inside the 48x48 pixel region. An edge margin of 32 pixels was grayed out in each patch. This was determined to provide the optimal context (see section 4.1 ``Does Context Help?''). 

\begin{figure}
\centering
\includegraphics[height=3.5 cm]{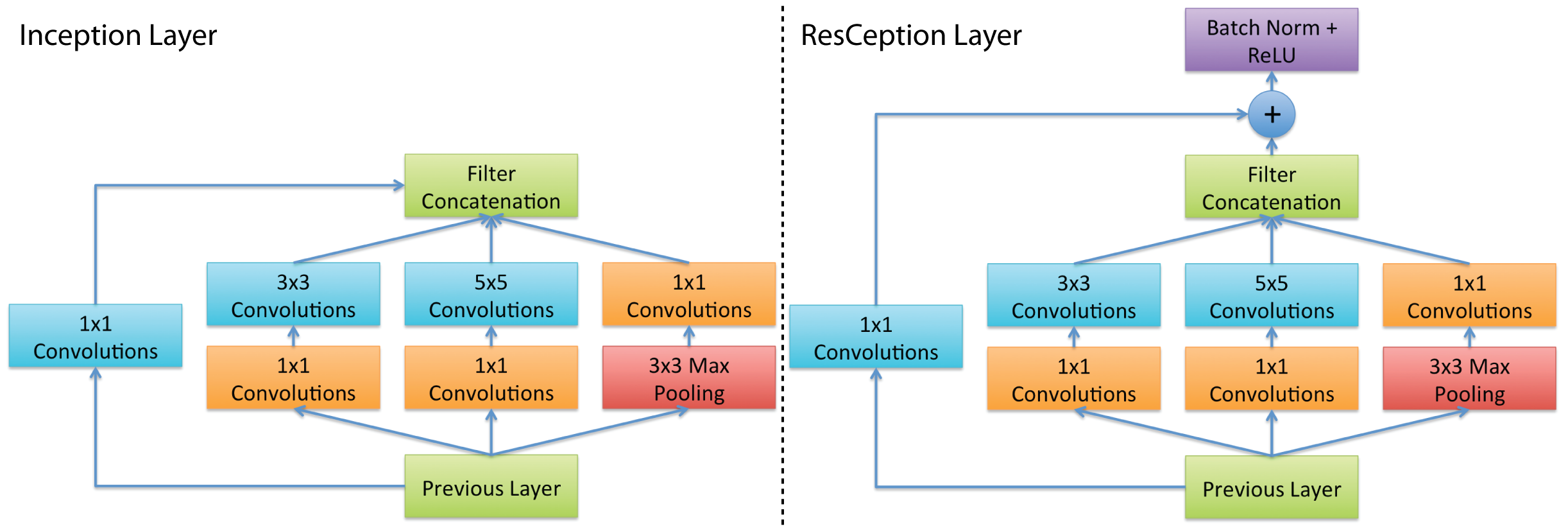}
\caption{{\it left} A standard Inception layer. {\it right} A ResCeption layer. The primary difference between it and Inception is that the 1x1 convolutions are used as a residual shortcut with projection. }
\label{fig:resception}
\end{figure}

We trained a few different networks for comparison. Since we want to include contextual information, larger more state-of-the-art networks are used. We used {\it AlexNet} \cite{AlexNet} as our smaller baseline and {\it GoogLeNet/Inception} with {\it batch normalization} \cite{GoogLeNet,BatchNorm}. We created a third network to synthesize {\it Residual Learning} \cite{ResNet} with Inception. We called this one {\it ResCeption}  (Fig.~\ref{fig:resception}). The ResCeption network is created by removing the 1x1 convolutions in each Inception layer and replacing them with a residual ``projection shortcut''. In section five, the advantage of doing this will become more apparent. \cite{Inceptionv4} published to {\it arXiv} at the time of this writing, is similar, but keeps the 1x1 convolution layers. These seem redundant with the residual shortcut which is why we removed them. The ResCeption version of GoogLeNet has about 5\% more operations than the Inception version, but interestingly runs about 5\% faster on our benchmark machine. All three networks were trained using {\it Caffe} \cite{Caffe} and stochastic gradient descent for 240k iterations with a mini batch size of 64. A polynomial rate decay policy was used with initial learning rate of 0.01 and power of 0.5. Momentum was 0.9 and weight decay 0.0002. The network input size was 224x224, so training images were randomly cropped to move the target a little around inside the 48x48 central region. Testing images were center-cropped. 

\begin{figure}
\centering
\includegraphics[height=4.25cm]{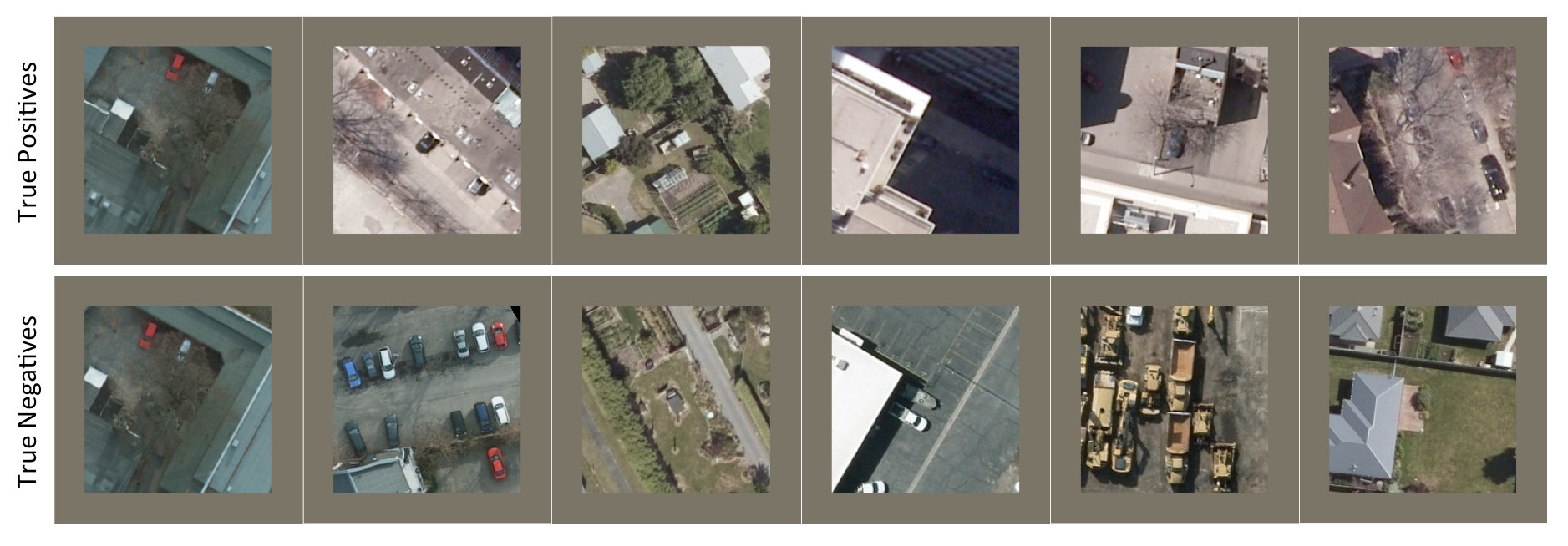}
\caption{ ({\it top row}) Test patches which the ResCeption network correctly classified as containing a car in the central region. Occlusions and visibility issues are commonly handled, but we note that they still appear to account for much of the error. 
({\it bottom row})  Patches that were correctly classified as not containing a car in the central region. The leftmost image is not a mistake. It has a tree in the center while the shifted version above it has a car concealed slightly underneath the tree. }
\label{fig:correct_examples}
\end{figure}

\setlength{\tabcolsep}{4pt}
\begin{table}
\begin{center}
\caption{ The percentage of test patches correctly classified for each deep model. The Non-Utah model was trained with non-Utah data (the other five sets) and then tested with Utah data to observe generalization to new datasets. }
\label{table:class_correct}
\begin{tabular}{ll}
\hline\noalign{\smallskip}
Model & Correct\\
\noalign{\smallskip}
\hline
\noalign{\smallskip}
AlexNet & 97.62\%\\
Inception & 99.12\%\\
ResCeption & 99.14\%\\
ResCeption Non-Utah & 98.89\%\\
\hline
\end{tabular}
\end{center}
\end{table}
\setlength{\tabcolsep}{1.4pt}
Table~\ref{table:class_correct} shows that Inception and ResCeption work noticeably better than AlexNet. However, all three seem to do pretty well.  Fig.~\ref{fig:correct_examples} shows examples of patches the ResCeption network got correct. 

\subsection{Does Context Help?}
\begin{figure}
\centering
\includegraphics[height=4.0cm]{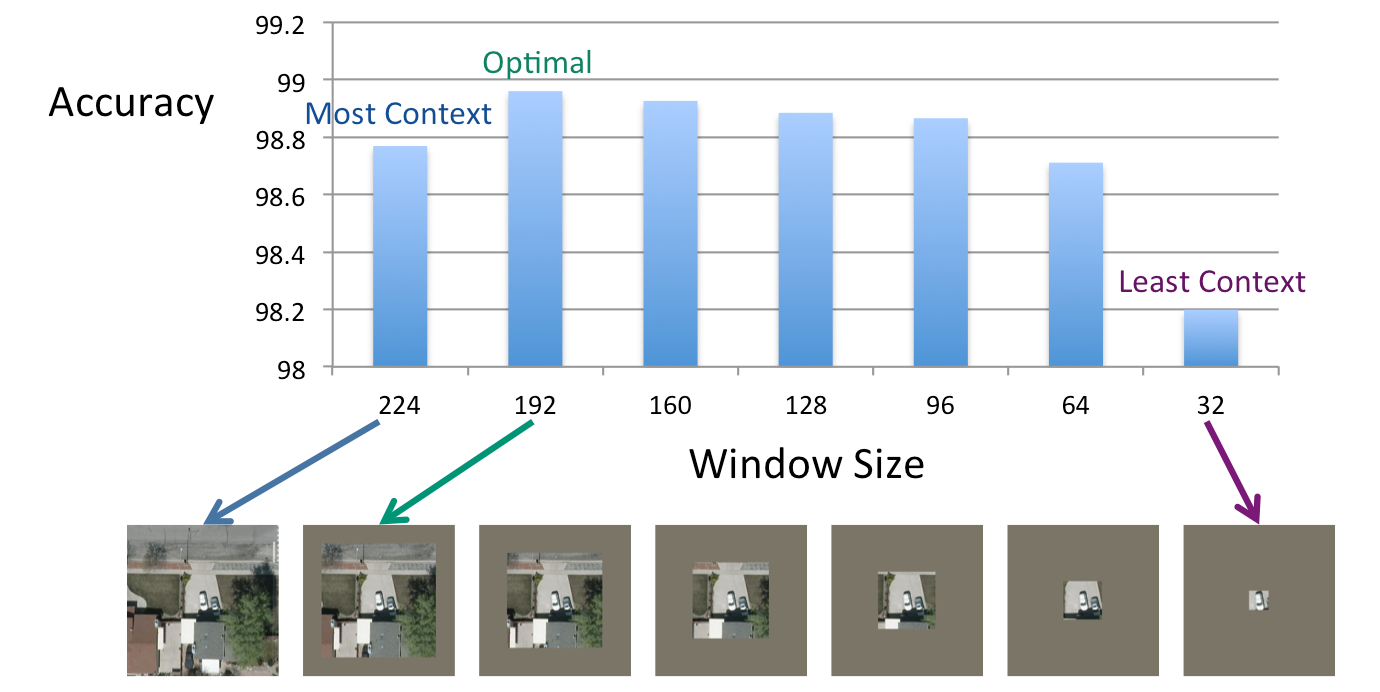}
\caption{The percentage of correct patches versus the amount of context present. As more context is included, accuracy improves. It appears optimal to cut out a small amount of context. }
\label{fig:context}
\end{figure}

We were interested to determining if context really helps classification results. To do this, we created sets where we masked out margins of the patches. By adjusting the size of the margin, we can determine performance changes as more or less of each patch is visible. We did this in increments of 32 pixels starting from the smallest possible region with only 32x32 pixels visible. Each training was done on GoogLeNet by fine-tuning the default version from Caffe. Fig.~\ref{fig:context} shows the results. Even the version with a small amount of context does well, but performance does increase monotonically until 192x192 pixels are visible. This suggests that most of the time, context is not all that important. Cars seem easy to classify. Context might only help in the 1\% or 2\% of difficult cases where strong occlusions and poor visibility make a determination difficult. That we can have too much context might be a result of too much irrelevant information or bad hints from objects that are too far from a car. 

\subsection{Detection}
Next we wanted to ascertain how well our trained network might perform on a real world task. One task of interest, is {\it target verification}. In this, another item such an {\it object tracker} \cite{Wu13} would have been assigned to track a target car. Our network would then be used to verify each frame to tell if the tracker was still tracking a car or if it had drifted to another target. A second more difficult task would involve localization and detection. This is the ability to find cars and determine where they are in a natural scene. The two tasks are almost equivalent in how we will test them. The biggest difference is how we score the results. 

For this task we used the trained ResCeption network since it had a slightly better result than Inception. Each of the 10 labeled 2048x2048 scene image were scanned with a stride of eight. At each stride location, 192x192 pixels were extracted and a 32 pixel margin was added to create a 224x224 patch. The softmax output was computed and taken to the power of 16 to create a wider gradient for the output around the extremes:
\begin{align}
p & = \left(o_{1} - o_{2} + 1\right)^{16}/2^{16}
\end{align}

\begin{figure}
\centering
\includegraphics[height=5.0cm]{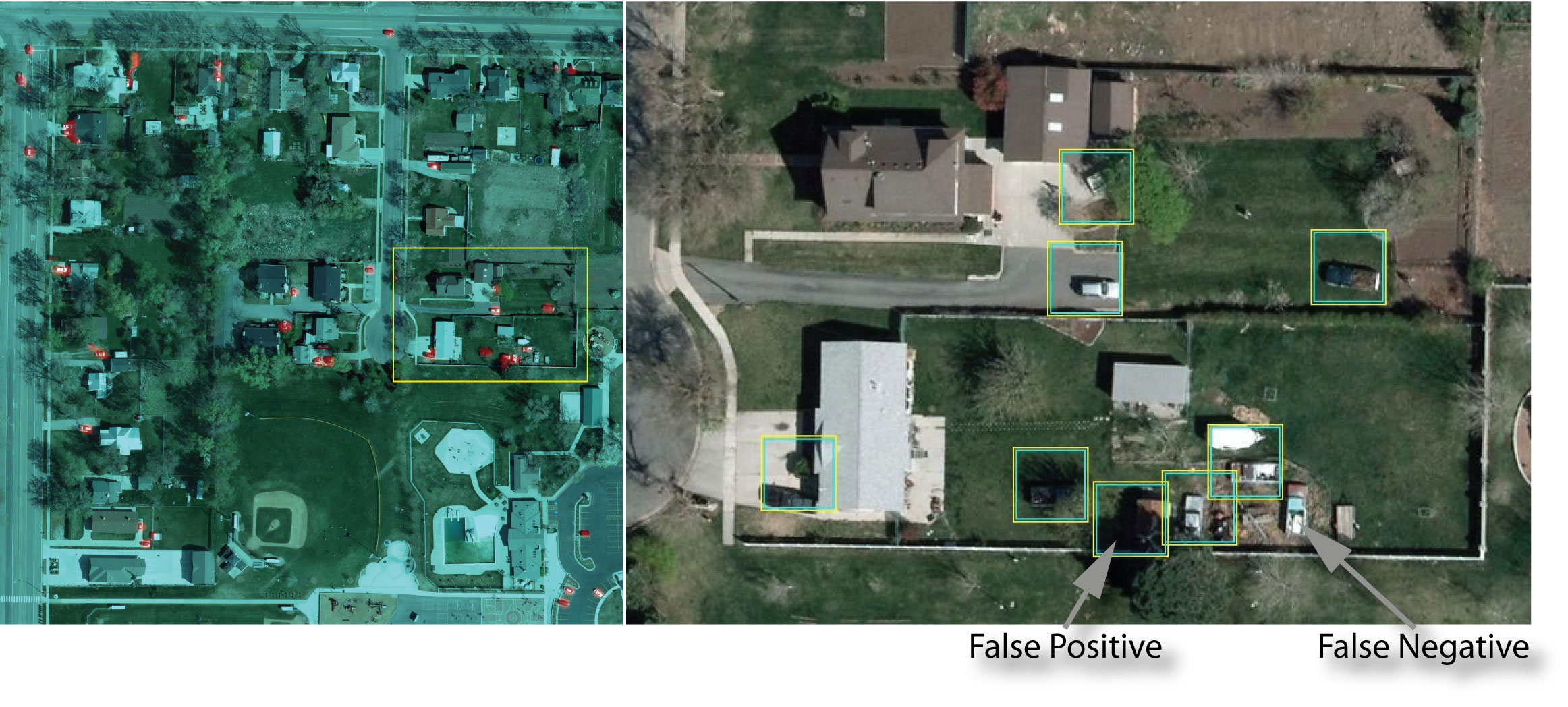}
\caption{ ({\it left}) A held out scene super imposed with the derived heat map colored in red. ({\it right}) A close up of one of the sections (highlighted in yellow on left) showing examples of detections and misses. The lower left car is considered detected since it is mostly inside the box. The false positive appears to be a shed. }
\label{fig:detections}
\end{figure}

This yielded a {\it heat map} with pixels $p$ created from softmax outputs {\it car}  $o_{1}$ and {\it not car} $o_{2}$. The value ranged from 0 to 1 with a strong upper skew. Location was determined using basic non-maximal suppression on the heat map (keeping it simple to establish a baseline). Bounding boxes were fixed at size 48 pixels which is the maximum length of a car. Boxes could overlap by as much as 20 pixels. Maximal locations were thresholded at 0.75 to avoid weak detections. These values were established on a special {\it validation scene}, not the hold out scenes. A car was labeled as a {\it detection} if at least half of its area was inside a bounding box. A car was a {\it false negative} if it was not at least half inside a box. A {\it false positive} occurred when a bounding box did not have at least half a car inside it. For the verification condition, {\it splits} (two or more detections on the same car) and {\it mergers} (two or more cars only covered by one box) did not count since these are not an effect that should impact its performance. For {\it detection}, a split yielded an extra false positive per extraneous detection. A merger was counted as a false negative for each extra undetected car. Fig.~\ref{fig:detections} shows examples of detections from our method. 

In table ~\ref{table:detect_correct}, we can see the results for both conditions.  Typically for car detections without explicit location constraints, precision/recall statistics range from 75\% to 85\% \cite{Chen13,VEDAI} but may reach 91\% if the problem is explicitly constrained to cars on pavement only \cite{Moranduzzo14,Holt09,Kamenetsky15}. This is not an exact comparison, but an F-score of 94.37\% over an unconstrained area of approximately 1 $km^2$ suggests we are doing relatively well. 

\setlength{\tabcolsep}{4pt}
\begin{table}
\begin{center}
\caption{ Verification and detection performance is shown for the ResCeption model. Count is the number of cars in the whole set. TP, FP and FN are true positive, false positive and false negative counts. F is the precision/recall related F-Score. Ideally the verification score should be similar to the patch correctness which was 99.14\%. So, we do incur some extra error. Detection error is higher since it includes splits and mergers. }
\label{table:detect_correct}
\begin{tabular}{llllllll}
\hline\noalign{\smallskip}
Condition & Count & TP & FP & FN & Precision & Recall & F\\
\noalign{\smallskip}
\hline
\noalign{\smallskip}
Verification &260& 253 & 9 & 7 & 96.56\% & 97.31\% & 96.93\%\\
Detection & 260 & 250 & 20 & 10 & 92.59\% & 96.15\% & 94.34\%\\
\hline
\end{tabular}
\end{center}
\end{table}
\setlength{\tabcolsep}{1.4pt}

\section{Counting}
The goal here was to create a one-look \cite{YOLO} counting network that would learn the combinatorial properties of cars in a scene. This is an idea that was previously described in \cite{Segui15} who counted objects in synthetic MNIST \cite{MNIST} scenes using a smaller five-layer network. The overhead scenes we wish to count from are too large for a single look since they can potentially span trillions of pixels. However, we may be able to use a very large stride and {\it count large patches at a time}.  Thus, the counting task is broken into two parts. The {\it first part} is learning to count by one-look over a large patch. The {\it second part} is creating a stride that counts objects in a scene one patch at a time. 

Training patches are sampled the same as for the classification task. However, the class for each patch is the number of cars in that patch. Very frequently, cars are split in half at the border of the patch. These cars are counted if the point annotation is at least 8 pixels into the visible region of the image. Thus, {\it a car must be mostly in the patch} in order to be counted. If a highly ambiguous object was in a patch, we did our best to determine whether it was or was not a car. This is different from the classification task were ambiguous objects could be left out. Here, they were too common as member objects that would incidentally appear in a patch even if not labeled. 

We trained AlexNet \cite{AlexNet}, Inception \cite{GoogLeNet,BatchNorm} and ResCeption networks which only differed from the classification problem in the number of outputs. Here, we used a softmax with 64 outputs. A regression output may also be reasonable, but the maximum number of cars in a scene is sufficiently small that a softmax is feasible. Also, we cover the entire countable interval. So there are no gaps that would necessitate using regression. In all the training patches, we never observed more than 61 cars. We rounded up to 64 in case we ever came upon a set that large and wanted to fine tune over it. 

We also trained a few new networks. The main idea for creating the ResCeption network was to allow us to stack Inception like layers much higher. This could give us the {\it lightweight} properties of GoogLeNet, but the ability to {\it go big} like with ResNet \cite{ResNet}. Here we have created a {\it double tall} GoogLeNet like network. This is done by repeating each ResCeption layer twice giving us 22 ResCeption layers rather than 11. It was unclear if we needed three error outputs like GoogLeNet, so we created two versions of our taller network. One double tall network has only one error output (o1) while another has three in the same fashion as GoogLeNet (o3). 

\begin{figure}
\centering
\includegraphics[height=2.5cm]{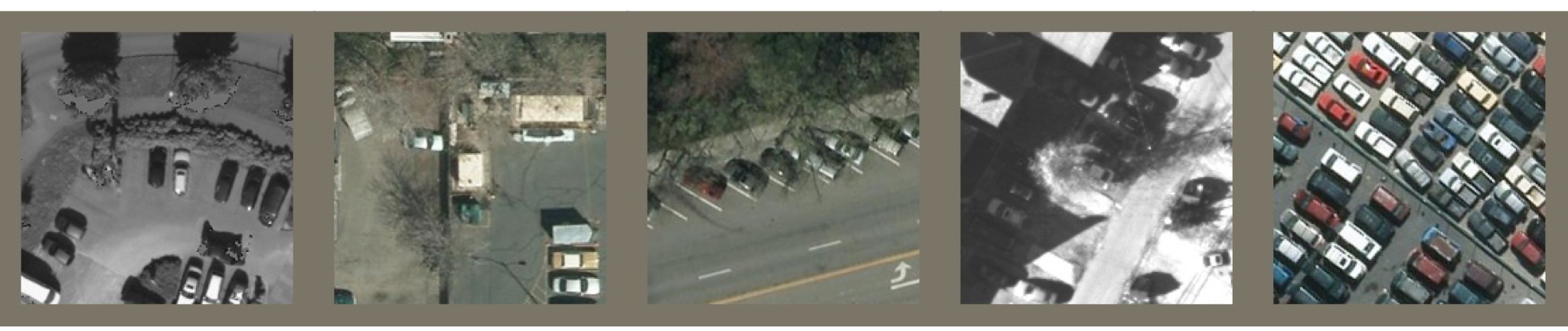}
\caption{ Examples of patches which were correctly counted by the ResCeption Taller o3 network. From left to right the correct number is 9, 3, 6, 13 and 47. Note that cars which are not mostly inside a patch are not counted. The center of the car must be at least 8 pixels inside the visible region. }
\label{fig:correct_patch_count}
\end{figure}

For the non-tall networks, training parameters were the same as with classification. For the double tall networks, the mini batch size was reduced to 43 to fit in memory and the training was extended to 360k iterations so that the same number of training exposures are used for both tall and non-tall networks. Table~\ref{table:count_patch} shows the results of error in counting on patch data. Examples of correct patch counts can be seen in Fig.~\ref{fig:correct_patch_count}. It's interesting to note that we can train a very tall GoogLeNet like network with only one error output. This suggests that {\it the residual component is doing its job}. 

\setlength{\tabcolsep}{4pt}
\begin{table}
\begin{center}
\caption{ The patch based counting error statistics.  The first data column is the percentage of test patches the network gets exactly correct. The next two columns show the percentage counted within 1 or 2 cars. MAE is the mean absolute error of count. RMSE is the root mean square error. The last column is the accuracy if we used the counting network as a proposal method. Thus, if we count zero cars, the region would be proposed to contain no cars. If the region contains at least one car, we propose that region has at least one car. The Taller ResCeption network with just one error output has the best metrics in three of the six columns. However, the improvement is somewhat modest. }
\label{table:count_patch}
\begin{tabular}{lllllll}
\hline\noalign{\smallskip}
Model & Correct & is +/- 1 & is +/- 2 & MAE & RMSE & Proposal Acc\\
\noalign{\smallskip}
\hline
\noalign{\smallskip}
AlexNet & 67.97\% & 95.69\% & 98.82\% & 0.527 & 1.192 & 95.32\%\\
Inception & 80.35\% & 95.89\% & 98.87\% & 0.257 & 0.665 & 97.79\%\\
ResCeption & 80.34\% & 95.95\% & 98.86\% & 0.255 & {\bf 0.657} & 97.69\%\\
ResCeption Taller o1 & {\bf 81.07}\% & {\bf 96.11\%} & 98.89\% & {\bf 0.248} & 0.676 & {\bf 97.84\%}\\
ResCeption Taller o3 & 80.82\% & 96.08\% & {\bf 98.95\%} & 0.250 & 0.665 & 97.83\%\\
\hline
\end{tabular}
\end{center}
\end{table}
\setlength{\tabcolsep}{1.4pt}

\subsection{Counting Scenes}
\begin{figure}
\centering
\includegraphics[height=5.0cm]{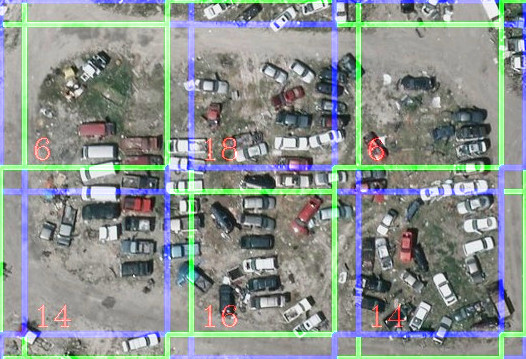}
\caption{A subsection of one of the held out scenes. It shows the stride used by the network as well as the number of cars it counted in each stride. Blue and green borders are used to help differentiate the region for each stride. One can see the overlapping region where the strides cross. 74 cars are counted among the six strides seen. The sub-scene contains 77 complete cars. Note that cars on their side are not counted since we are only concerned with cars that are mobile. }
\label{fig:stride}
\end{figure}

One of the primary difficulties in moving from counting cars in patches to counting cars in a scene using a large stride is that if we do not have overlap between strides, cars may be cut in half and counted twice or not at all. Since cars have to be mostly inside a patch, some overlap would be ideal. Thus, by requiring that cars are mostly inside the patch, we have created a remedy to the splitting and counting twice problem, but have increased the likelihood of not counting a split car at all. In this case, since we never mark the location of cars, there is no perfect solution for this source of error. We cannot eliminate cars as we count because we do not localize them. However, {\it we can adjust the stride to minimize the likelihood of this error}. We used the special validation scene with 628 cars and adjusted the stride until error was as low as possible. This resulted in a stride of 167. We would then use this stride in any other counting scene. An example of a strided count can be seen in Fig.~\ref{fig:stride}. To allow any stride and make sure each section was counted, we padded the scene with zeros so that the center of the first patch starts at (0,0) in the image.

We tested counting on a held out set of 20 2048x2048 scenes. The number of cars in each scene varied between 881 and 10 with a mean of 173. Each held out scene was from the Utah AGRC image set. However, we selected geolocations which were unique. All Utah patch data came from the Salt Lake City metropolitan area. The held out data came from locations outside of that metro. These included some unusual and difficult locations such as {\it mountain forest}, a {\it water park} and an {\it auto junkyard}. Many of the scenes have dense parking lots, but many do not. We included industrial, commercial and residential zones of different densities. The error results can be seen in Table~\ref{table:count_scene_s1}. It is interesting to note that while the double tall ResCeption network with one output dominates the patch results, {\it the double tall ResCeption network with three outputs dominates the scene results}. This may be related to the lower RMSE and +/- 2 error for the three-output network. The network may be less prone to outlier errors. The two extra error inputs may act to regulate it.  

\setlength{\tabcolsep}{4pt}
\begin{table}
\begin{center}
\caption{Error for counting over all 20 held out scenes. Mean absolute error is expressed as a {\it percentage of cars} over or under counted. This is taken as a percentage since the mean absolute error (MAE) by count highly correlates with the number of cars in a scene (r $>$ 0.80 for all models). RMSE is the root mean square of the percent errors. The maximum error shows what the largest error was on any of the 20 scenes. Cars in ME is how many cars were in the scene with the highest error. Scenes with smaller numbers of cars can bump error up strongly since missing one or two cars is a larger proportion of the count. Finally, we count how many cars are in the entire set of 20 scenes. The total error shows us how far off from the total count we are when we sum up all the results. So for instance, there are a total of 3456 cars in all 20 scenes. The taller ResCeption network with three outputs counts 3472 cars over all the scenes. Its total error is 0.46\%. A low total error suggests that error between scenes is more random and less biased since it centers around zero. This seems to contradict the correlation between error and size mentioned earlier. This may come about if there is a bias within scenes, but not between scenes (i.e. some types of scenes tend to be over counted and others tend to be under counted and this cancels out when we sum the scene counts). }
\label{table:count_scene_s1}
\begin{tabular}{llllll}
\hline\noalign{\smallskip}
Model & MAE & RMSE & Max Error & Cars in ME & Total Error\\
\noalign{\smallskip}
\hline
\noalign{\smallskip}
AlexNet & 8.46\% & 11.64\% & 27.27\% & 22 & 3.30\%\\
Inception & 6.50\% & 8.05\% & 17.65\% & 51 & 0.84\%\\
ResCeption & {\bf 5.78\%} & 8.09\% & 18.18\% & 22 & 1.22\%\\
ResCeption Taller o1 & 6.44\% & 8.09\% & 18.18\% & 22 & 1.19\%\\
ResCeption Taller o3 & 6.14\% & {\bf 7.57\%} & {\bf 15.69\%} & 51 & {\bf0.46\%}\\
\hline
\end{tabular}
\end{center}
\end{table}
\setlength{\tabcolsep}{1.4pt}

As a second experiment, we attempted to reduce the error caused by double counting or splitting by {\it averaging different counts} over the same scene. Each count has a different offset. So we have the original stride which starts at (0,0) and three new ones which start at (0,4), (4,0) and (4,4). Ideally, these slight offsets should not split or double count the same cars. The results can be seen in Table~\ref{table:count_scene_s4}. Of the twenty scene error metrics we consider, 19 are reduced by averaging over several strides. 

\setlength{\tabcolsep}{4pt}
\begin{table}
\begin{center}
\caption{ This is similar to Table~\ref{table:count_scene_s1} but here we show the mean result from four different slightly offset strides. In 19 of the 20 error statistics, this method improves results over  ~\ref{table:count_scene_s1}. With some random error due to double counting removed, the three output taller ResCeption model is clearly superior.  }
\label{table:count_scene_s4}
\begin{tabular}{llllll}
\hline\noalign{\smallskip}
Model & MAE & RMSE & Max Error & Cars in ME & Total Error\\
\noalign{\smallskip}
\hline
\noalign{\smallskip}
AlexNet & 8.40\% & 10.53\% & 21.59\% & 22 & 3.02\%\\
Inception & 6.46\% & 7.86\% & 15.69\% & 51 & 0.75\%\\
ResCeption & 5.35\% & 7.17\% & 14.77\% & 22 & 1.12\%\\
ResCeption Taller o1 & 5.85\% & 6.95\% & 13.24\% & 51 & 1.22\%\\
ResCeption Taller o3 & {\bf 5.15\%} & {\bf 6.70\%} & {\bf 12.75\%} & 51 & {\bf 0.20\%}\\
\hline
\end{tabular}
\end{center}
\end{table}
\setlength{\tabcolsep}{1.4pt}

A comparison with other car counting methods with available data can be seen in Table~\ref{table:count_method}. Mean accuracy is comparable to interactive density estimation \cite{Arteta14}. However, our method is completely automatic. Scenes do not need any special interactive configuration. Another item of interest is that we make no explicit assumption about car location. \cite{Kamenetsky15} uses a pavement locator to help reduce false positives. In our case, cars are counted even if they are on someone's lawn. The deep learner may potentially ingest and understand context, so it is conceivable that it may be biased against counting a car in water or on a building top.

\setlength{\tabcolsep}{4pt}
\begin{table}
\begin{center}
\caption{Reported errors for two recent car counting methods are shown compared with the error from our best model's results. The first column indicates if the method is completely automatic. The second column tells us if we do not have any location restrictions such as only counting cars on roads or scenes that have been corrected. The Images column is how many scene images are in the test set. Total cars over all scenes is shown after that as well as how many cars were counted in total. The mean absolute error is given over all the test scenes. For the SIFT/SVM method, one single scene accounts for much of the error. This is not an apples-to-apples comparison, but it does give a general idea of performance given the strengths of our approach.}
\label{table:count_method}
\begin{tabular}{llllllll}
\hline\noalign{\smallskip}
Method & Auto. & No Loc. & Images & Tot. Cars & Counted & MAE & Tot. Error\\
\noalign{\smallskip}
\hline
\noalign{\smallskip}
Density \cite{Arteta14} & No & No & 1 & 230 & 220 & 4.35\% & 4.35\%\\
SIFT/SVM \cite{Moranduzzo14} & Yes & No & 5 & 119 & 132 & 36.74\% & 9.85\%\\
Deep Learn & Yes & Yes & 20 & 3456 & 3463 & 5.15\% & 0.19\%\\
\hline
\end{tabular}
\end{center}
\end{table}
\setlength{\tabcolsep}{1.4pt}

\subsection{Counting Efficiency}
In addition to accuracy, we wanted to measure the efficiency of our solution. We are using larger networks, but we are also using a very large stride. The cost of running GoogLeNet is 30k ops per pixel at 224x224. With a stride of 167 on a scene, the cost increases to 54k ops per pixel over the scene. By modern standards, this is not an outrageous cost. By comparison, a very small, single-pixel strided CNN would require at least a million ops per pixel over a scene. Table~\ref{table:count_time} shows the time of running the different counting network over a scene.  

\setlength{\tabcolsep}{4pt}
\begin{table}
\begin{center}
\caption{Performance results taken for our models running on {\it Caffe} on a single {\it Nvidia GeForce Titan X} based GPU. The number of batches required to run a full scene gives us an idea of the extra overhead from running larger models. The time is how many seconds it takes to run a single 2048x2048 scene. This yields a rate in {\it fps}. Finally we can see how many $km^2$ we can scan per second using the method. }
\label{table:count_time}
\begin{tabular}{llllll}
\hline\noalign{\smallskip}
Model & Batches & Time & FPS & $km^2$ PS\\
\noalign{\smallskip}
\hline
\noalign{\smallskip}
AlexNet & 1 & 0.087 & 11.486 & 1.084\\
Inception & 2 & 0.366 & 2.731 & 0.258\\
ResCeption & 2 & 0.344 & 2.906 & 0.274\\
ResCeption Taller o1 & 4 & 0.748 & 1.337 & 0.126\\
ResCeption Taller o3 & 4 & 0.773 & 1.294 & 0.122\\
\hline
\end{tabular}
\end{center}
\end{table}
\setlength{\tabcolsep}{1.4pt}

The AlexNet version will count cars at a rate of 1 $km^2$ per second. A company such as {\it Digital Globe} which produced satellite data at the rate of 680,000 $km^2$ per day in 2014 would theoretically be able to count the cars in all that data {\it online} with 8 GPUs. Indeed, another comparison would be to \cite{OrbitalInsight16}. As their solution is proprietary, comparison data is difficult to come by. They have claimed that they can count cars in {\it 4 trillion pixels worth of images in 48 hours} using a cloud-based solution. Their approach is to label each pixel using a deep network \cite{Brust15} for the pixel's ``car-ness''. Assuming image data is supplied to the GPU just in time, our AlexNet based solution would be able to count that many pixels in {\it 23 hours using one single GPU}. AlexNet has 8.46\% mean absolute error, but if one is just analyzing trends such as number of customers at a shopping center, this is probably accurate enough.

\section{Conclusion}

We have created a large and difficult dataset of cars overhead that we have used to classify, detect and count cars. Our classification results are quite excellent and our detection results appear to be better than even those of methods that constrain the location of cars. Out counting method appears to be very efficient and yields results similar to methods which are scene constrained or need to be fine tuned to process scenes other than the ones used in training.  

\subsubsection{Acknowledgments.}

This work was funded from the NA-22 project at Lawrence Livermore National Laboratory's Global Security directorate. Thanks to ISPRS, DGPF and BSF Swissphoto for permission to use their data. 

\bibliography{egbib}

\begin{thebibliography}{10}

\bibitem{OrbitalInsight16}
Crawford, J.:
\newblock Beyond supply and demand: Making the invisible hand visible.
\newblock In: Re-Work Deep Learning Summit, San Francisco (January 2016)

\bibitem{OIRDS}
Tanner, F., Colder, B., Pullen, C., Heagy, D., Eppolito, M., Carlan, V.,
  Oertel, C., Sallee, P.:
\newblock Overhead imagery research data set: an annotated data library and
  tools to aid in the developement of computer vision algorithms.
\newblock In: IEEE Applied Imagery Pattern Recognition Workshop. (2009)

\bibitem{VEDAI}
Razakarivony, S., Jurie, F.:
\newblock Vehicle detection in aerial imegery : A small target detection
  benchmark.
\newblock Journal of Visual Communication and Image Representation (December
  2015) $<$hal-01122605v2$>$.

\bibitem{UtahData}
{Utah Automated Geographic Reference Center (AGRC)}:
\newblock {Utah 2012 HRO} 6 inch orthophotography data.
  http://gis.utah.gov/data/aerial-photography/.

\bibitem{TorontoData}
{International Society for Photogrammetry and Remote Sensing (ISPRS)}:
\newblock {WG3 Toronto} overhead data. http://www2.isprs.org/commissions/comm3
  /wg4/tests.html.

\bibitem{SelwynData}
{Land Information New Zealand (LINZ)}:
\newblock Selwyn 0.125m urban aerial photos index tiles {(2012-13)}.
  https://data.linz.govt.nz/layer/1926-selwyn-0125m-urban-aerial-photos-2012-13/.

\bibitem{PotsdamData}
{International Society for Photogrammetry and Remote Sensing (ISPRS) and BSF
  Swissphoto}:
\newblock {WG3 Potsdam} overhead data. http://www2.isprs.org/commissions/comm3
  /wg4/tests.html.

\bibitem{VaihingenData}
{International Society for Photogrammetry and Remote Sensing (ISPRS) and the
  German Society of Photogrammetry, Remote Sensing and Geoinformation (DGPF)}:
\newblock {WG3 Vaihingen} overhead data.
  http://www2.isprs.org/commissions/comm3 /wg4/tests.html.

\bibitem{ColumbusData}
{United States Air Force Research Lab (AFRL)}:
\newblock Columbus surrogate unmanned aerial vehicle {(CSUAV)} dataset.
  https://www.sdms.afrl.af.mil/index.php?collection=csuav.

\bibitem{Chen13}
Chen, X., Xiang, S., Liu, C.L., Pan, C.H.:
\newblock Vehicle detection in satellite images by parallel deep convolutional
  neural networks.
\newblock In: Second IAPR Asian Conference on Pattern Recognition. (2013)

\bibitem{Moranduzzo14}
Moranduzzo, T., Melgani, F.:
\newblock Automatic car counting method for unmanned aerial vehicle images.
\newblock IEEE Transactions on Geoscience and Remote Sensing \textbf{52}(3)
  (2014)  1635--1647

\bibitem{Holt09}
Holt, A.C., Seto, E.Y.W., Rivard, T., Gong, P.:
\newblock Object-based detection and classification of vehicles from
  high-resolution aerial photography.
\newblock Photogrammetric Engineering and Remote Sensing \textbf{75}(7) (2009)
  871--880

\bibitem{Kamenetsky15}
Kamenetsky, D., Sherrah, J.:
\newblock Aerial car detection and urban understanding.
\newblock In: IEEE Conference on Digital Image Computing: Techniques and
  Applications {(DICTA)}. (2015)

\bibitem{Zhang15}
Zhang, C., Li, H., Wang, X., Yang, X.:
\newblock Cross-scene crowd counting via deep convolutional neural networks.
\newblock In: CVPR. (2015)

\bibitem{Arteta14}
Arteta, C., Lempitsky, V., Noble, J.A., Zisserman, A.:
\newblock Interactive object counting.
\newblock In: ECCV. (2014)

\bibitem{Lempitsky10}
Lempitsky, V., Zisserman, A.:
\newblock Learning to count objects in images.
\newblock In: NIPS. (2010)

\bibitem{French15}
French, G., Fisher, M.H., Mackiewicz, M., Needle, C.L.:
\newblock Convolutional neural network for counting fish in fisheries
  surveillance video.
\newblock In: BMVC. (2015)

\bibitem{Idrees13}
Idrees, H., Saleemi, I., Seibert, C., Shah, M.:
\newblock Multi-source multi-scale counting in extremely dense crowd images.
\newblock In: CVPR. (2013)

\bibitem{YOLO}
Redmon, J., Divvala, S., Girshick, R., Farhadi, A.:
\newblock You only look once: Unified, real-time object detection.
\newblock In: arXiv preprint arXiv:1506.02640. (2015)

\bibitem{Segui15}
Segue, S., Pujol, O., Vitria, J.:
\newblock Learning to count with deep object features.
\newblock In: CVPR. (2015)

\bibitem{Wang15}
Wang, C., Zhang, H., Yang, L., Liu, S., Cao, X.:
\newblock Deep people counting in extremely dense crowds.
\newblock In: Proceedings of the 23rd Annual ACM Conference on Multimedia.
  (2015)

\bibitem{Kelly09}
Kelly, J., Sukhatme, G.S.:
\newblock Visual-inertial simultaneous localization, mapping and
  sensor-to-sensor self-calibration.
\newblock In: CIRA. (2009)

\bibitem{AlexNet}
Krizhevsky, A., Sutskever, I., Hinton, G.E.:
\newblock {ImageNet} classification with deep convolutional neural networks.
\newblock In: NIPS. (2013)

\bibitem{GoogLeNet}
Szegedy, C., Liu, W., Jia, Y., Sermanet, P., Reed, S., Anguelov, D., Erhan, D.,
  Vanhoucke, V., Rabinovich, A.:
\newblock Going deeper with convolutions.
\newblock In: CVPR. (2015)

\bibitem{BatchNorm}
Ioffe, S., Szegedy, C.:
\newblock Batch normalization: Accelerating deep network training by reducing
  internal covariate shift.
\newblock In: ICML. (2015)

\bibitem{ResNet}
He, K., Zhang, X., Ren, S., Sun, J.:
\newblock Deep residual learning for image recognition.
\newblock In: arXiv preprint arXiv:1512.03385. (2015)

\bibitem{Inceptionv4}
Christian~Szegedy, Sergey~Ioffe, V.V.:
\newblock Inception-v4, inception-resnet and the impact of residual connections
  on learning.
\newblock In: arXiv:1602.07261. (2016)

\bibitem{Caffe}
Jia, Y., Shelhamer, E., Donahue, J., Karayev, S., Long, J., Girshick, R.,
  Guadarrama, S., Darrell, T.:
\newblock Caffe: Convolutional architecture for fast feature embedding.
\newblock arXiv preprint arXiv:1408.5093 (2014)

\bibitem{Wu13}
Wu, Y., Yang, M.H., Lim, J.:
\newblock Online object tracking: A benchmark.
\newblock In: CVPR. (2013)

\bibitem{MNIST}
LeCun, Y., Cortes, C., Burges, C.J.:
\newblock The {MNIST} database of handwritten digits.
  http://yann.lecun.com/exdb/mnist/.

\bibitem{Brust15}
Brust, C.A., Sickert, S., Simon, M., Rodner, E., Denzler, J.:
\newblock Efficient convolutional patch networks for scene understanding.
\newblock In: CVPR Scene Understanding Workshop. (2015)

\end{thebibliography}
\bibliographystyle{splncs}
\end{document}